\documentclass[11pt,a4paper]{article}
\usepackage[hyperref]{eacl2021}
\usepackage{times}
\usepackage{latexsym}

\usepackage{amsmath}
\usepackage{graphicx}
\usepackage{booktabs}
\usepackage{array}
\usepackage{microtype}
\usepackage{verbatim}

\aclfinalcopy 
\def\aclpaperid{1314} 

\title{Recipes for Adapting Pre-trained Monolingual and Multilingual Models to Machine Translation}

\author{Asa Cooper Stickland$^\clubsuit$ \\ \\ \And  Xian Li$^\spadesuit$ \\ \\ $^\clubsuit$ University of Edinburgh, $^\spadesuit$ Facebook AI \\
\texttt{a.cooper.stickland@ed.ac.uk, \{xianl,ghazvini\}@fb.com} \\ \\ \And Marjan Ghazvininejad$^\spadesuit$  \\
 }

\date{}

\begin{document}
\maketitle
\begin{abstract}
There has been recent success in pre-training on monolingual data and fine-tuning on Machine Translation (MT), but it remains unclear how to best leverage a pre-trained model for a given MT task.
This paper investigates the benefits and drawbacks of freezing parameters, and adding new ones, when fine-tuning a pre-trained model on MT.
We focus on 1) Fine-tuning a model trained only on English monolingual data, BART\@. 2) Fine-tuning a model trained on monolingual data from 25 languages, mBART\@. For BART we get the best performance by freezing most of the model parameters, and adding extra positional embeddings. For mBART we match or outperform the performance of naive fine-tuning for most language pairs with the encoder, and most of the decoder, frozen. The encoder-decoder attention parameters are most important to fine-tune. When constraining ourselves to an out-of-domain training set for Vietnamese to English we see the largest improvements over the fine-tuning baseline.
\end{abstract}

\section{Introduction}

\begin{figure}[t]
\vskip 0.2in
\begin{center}
\centerline{\includegraphics[width=3in]{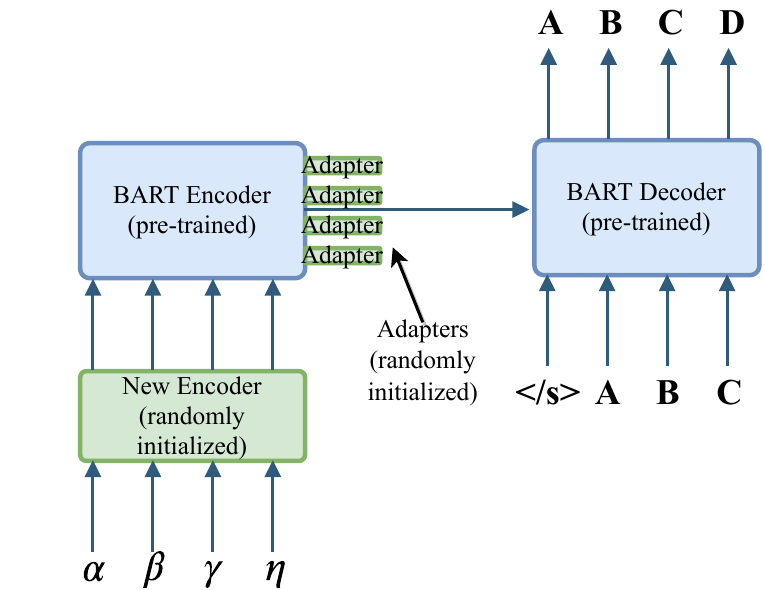}}
\caption{Schematic diagram showing the components of our system for adapting BART to MT. We learn a new encoder that takes as input the source language, with a potentially different vocabulary to the original BART system. We freeze most BART parameters (frozen model components are shown in blue).}
\label{fig:bart}
\end{center}
\vskip -0.2in
\end{figure}

\begin{figure}[t]
\vskip 0.2in
\begin{center}
\centerline{\includegraphics[width=3in]{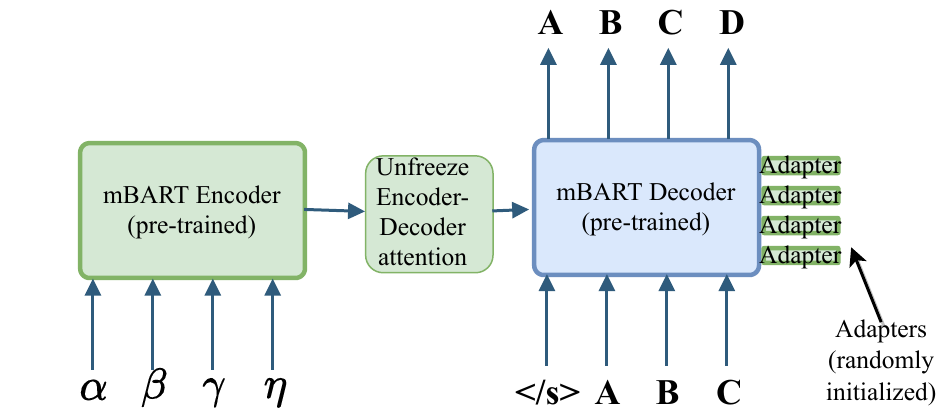}}
\caption{Schematic diagram showing one method of adapting mBART to MT, unfreezing the encoder and encoder-decoder attention, and adding adapters in the decoder. Model components colored blue are not updated during fine-tuning.}
\label{fig:mbart}
\end{center}
\vskip -0.2in
\end{figure}
Machine Translation (MT) has recently seen significant advances, with improvements in modeling, especially since the advent of neural models \cite{seq, bahdanau}, and the  availability of large parallel corpora for training such systems \cite{smith-etal-2013-dirt, kocmi-bojar-2017-curriculum, tiedemann-2012-parallel}. However, often standard neural systems do not perform well on \textit{low-resource} language pairs \cite{six}, especially when the language pairs are only distantly related. Since these languages are spoken by a large fraction of the world's population, reducing the gap in performance between high and low-resource MT could have a large impact.
  
An explosion of interest in large-scale pre-training in Natural Language Processing has led to increased performance on smaller datasets, by simple \textit{fine-tuning} of large pre-trained models on downstream tasks. 
The typical approach is to train a large model on text from the web (for example English Wikipedia), with a common objective predicting masked out tokens using the unmasked context. For Natural Language Generation (for example summarization of text), performance can be improved by pre-training a sequence-to-sequence model~\cite{song2019mass,bart}. 

However previous work has shown that on NLP tasks such as Natural Language Inference, the relative performance of fine-tuning vs. keeping the pre-trained model frozen depends on the similarity of the pre-training and downstream tasks \cite{peters-freezing}. We observe empirically that simple fine-tuning of a monolingual model for MT can result in worse performance than training from scratch (e.g.\ Table~\ref{tab:freeze-ablatate}). For MT the more common monolingual (usually only English) pre-training~\cite{peters2018deep,radford2018gpt,devlin2018bert,yang2019xlnet,liu2019roberta} may be inadequate since the input or output domain for the downstream task will be a non-English language.

\textit{Multilingual} pre-training offers a solution, by modifying the pre-training objective to include many languages. Using a multilingual pre-trained model for MT gives good performance, especially on lower-resource language directions \cite{mbart}. However it is challenging to balance the training data so that higher-resource languages do not overwhelm lower-resource ones \cite{massive, xlmr}. For a particular language it may be hard to source monolingual data, or it may be simply not included in training. 

We also consider multilingual MT (training on many language pairs and sharing all or most model parameters) as a downstream task. Sharing 'knowledge' across language directions can improve performance on low-resource language pairs by transfer from other pairs included in training.
Previous work observed problems of performance degradation, often on high-resource languages, due to interference and constrained capacity \cite{johnson2017google, tan-etal-2019-multilingual}. And when initialising from a pre-trained model, we want to avoid `catastrophic forgetting', where by fine-tuning on a particular language pair we lose the knowledge about another language pair that is stored in the model weights. 

Previous work has explored how to improve on simple fine-tuning, by freezing pre-trained model parameters \cite{peters-freezing, houls} and using lightweight `adapter modules' \cite{houls, pals} which are inserted between the layers of the pre-trained network.
We aim to explore and improve on these approaches for both bilingual and multilingual MT (in contrast to previous work largely focusing on text classification).  We explore freezing different subsections of the pre-trained model.
We expect freezing to be particularly useful when the parallel data is of low quality, in which case naive fine-tuning may, for example, over-specify the pre-trained model to a particular domain.

Our main contributions are:
\begin{itemize}
    \item A novel fine-tuning approach, similiar to \citet{bart} but with adapter modules in the encoder of the pre-trained sequence-to-sequence model and combining both learnable, and fixed sinusoidal, positional embeddings in the input module (see sections~\ref{sec:im} and~\ref{sec:extra}) that feeds into the pre-trained encoder.
    \item Extensive experiments with fine-tuning a multilingual pre-trained model for MT, showing the benefits and drawbacks of freezing various parameters. We find we should freeze the decoder but unfreeze the encoder-decoder attention when fine-tuning on \textit{Xx} $\rightarrow$ \textit{En} data, and in the other direction we should freeze the encoder but unfreeze the entire decoder (section~\ref{sec:what}). We find monolingual models benefit more from freezing parameters than multilingual models (section~\ref{sec:mbart-res}). 
    \item Results on fine-tuning a multilingual pre-trained model for multilingual MT showing that freezing parameters improves performance on some, mostly distantly related, language directions (section~\ref{sec:multi}). 

\end{itemize}
\section{Background and Related Work}
\label{sec:related}
\paragraph{BART and mBART}~
We briefly describe the pre-trained models we focus on in this work. In order to perform machine translation with the minimum of modifications to the pre-trained model, we prefer models that can perform conditional sequence generation. We concentrate on the BART (Bidirectional and Auto-Regressive Transformer) model \cite{bart} and the multilingual BART \citep[mBART;][]{mbart} model. BART and mBART are sequence-to-sequence models with the standard transformer-based neural machine translation architecture, i.e. an encoder and autoregressive decoder. The pre-training task they are trained on is reconstructing a document from a noisy version of that document (so called `de-noising autoencoder'). Examples of noise added to the training data include randomly shuffling the order of the original sentences, randomly changing the start position of the document, and using a masking scheme where arbitrary length spans of text are replaced with a single mask token. BART and mBART are trained entirely on monolingual data from the web, with English data for BART and data from 25 different languages for mBART\@. 

BART and mBART have almost identical architectures, with 12 encoder layers and 12 decoder layers with model dimension of 1024 and 16 attention heads. BART has a vocabulary of approximately 40k and $\sim$ 406M parameters, whereas mBART has a larger vocabulary of size 250k and $\sim$ 610M parameters.

\paragraph{Pre-trained Models for MT}~
There has been much recent progress in pre-training for NLP applications~\cite{peters2018deep,radford2018gpt,devlin2018bert,yang2019xlnet,liu2019roberta}, with the most relevant for our work focusing on text generation~\cite{radford2019language,song2019mass,dong2019unified,raffel2019exploring,bart}
Specifically for MT,
\newcite{ramachandran2017unsupervised} proposed pre-training the encoder-decoder modules as two separate language models, and \newcite{yang2019towards,zhu2020incorporating} explored approaches incorporating BERT model weights into the usual seq-to-seq architecture.

\paragraph{Multilingual MT} 

\textit{Multilingual translation}~\cite{firat2016multi,viegas2016google,aharoni-etal-2019-massively,massive} aims to jointly train one translation model that translates multiple language directions, and shares representations to improve the translation performance on low-resource languages~\cite{gu-etal-2018-universal}. Our freezing approach is similar in spirit to \citet{sachan-neubig-2018-parameter} who investigate which parameters are most useful to share for multilingual MT with transformer models. We start from a multilingual pre-trained model, and decide between sharing or freezing parameters.  
\paragraph{Transfer Learning for MT} 
\textit{Transfer learning} hopes to leverage a related task to perform well on a target task, for example by initialising the model weights from those resulting from training on a related task. For MT various approaches have been explored, with a common method training on high-resource language(s) and fine-tuning on a low-resource language \cite{neubig-hu-2018-rapid}. 

Closely related to our work is that of \citet{bapna-firat-2019-simple}, who introduce freezing and adapters (extra parameters inserted within the transformer) for domain adaption in MT. They take an MT model trained on a large parallel corpus, and fine-tune in a different domain (e.g.\ legal text). We differ in that we start from a pre-trained model that has not been trained on parallel text, and study adapting it to MT. Approaches based on freezing various model components have also been proposed \cite{thompson-etal-2018-freezing, zoph-etal-2016-transfer}, but have focused on RNN models pre-trained with parallel data, not transformer models pre-trained on monolingual data.

\section{Methods}

Because BART has been trained on only English input, we need to use different techniques when fine-tuning BART and mBART for MT, with a schematic overview shown in Figure~\ref{fig:bart} and Figure~\ref{fig:mbart}.
BART and mBART are standard sequence-to-sequence models, where an \textit{encoder} consumes a sequence of source-side tokens, and a \textit{decoder} acts as a conditional language model, generating target tokens given a source sequence. Intuitively, we want the encoder and decoder to be performing roughly the same tasks during fine-tuning as they were during pre-training. For BART\, this means the input to the encoder should be similar to (embedding vectors of) noisy English text. Therefore when training on say, Vietnamese to English, we first transform the Vietnamese source sentence into a representation useful for BART. We introduce new parameters (the `Input Module') that consume the source sentence and produce hidden vectors we can feed into the BART encoder. We describe the Input Module architecture in section~\ref{sec:im}. 

mBART can be fine-tuned without modification since during pre-training it saw the languages it will be fine-tuned on. To increase flexibility when freezing parts of the network, we optionally add extra parameters to both BART and mBART, described in section~\ref{sec:adapter}.

\subsection{Input Module Architecture}
\label{sec:im}

We refer to the network that takes in the source language text and outputs hidden vectors useful for BART as an `Input Module' or $\mathrm{IM}(\cdot)$. To improve performance on low-resource MT, we use smaller token embedding vectors on the source side of size $d_\mathrm{s} = 512$, whereas BART uses hidden vectors of size $d_\mathrm{BART}=1024$.  The full network is as follows, with $\{\mathbf{e}_t\}_{t=0}^{l}$ token embeddings for a source sentence with $l$ tokens,
\begin{equation}
\mathrm{BART}(\mathrm{IM}(\{\mathbf{e}_t\}_{t=0}^{l})),
\end{equation}
where $\mathrm{BART(\cdot)}$ is the full BART encoder-decoder model. Where we would normally input token embeddings to the BART model we use the outputs of the Input Module. The $t$-th element of $\mathrm{IM}(\{\mathbf{e}_t\}_{t=0}^{l})$ as follows:
\begin{equation}
    \alpha \mathrm{LN}(\mathbf{W}\mathrm{Transformer}(\{\mathbf{e}_t\}_{t=0}^{l})_t)
\end{equation}

 and where LN($\cdot$) is layer-norm, $\mathbf{W}$ is a matrix projecting up from $d_\mathrm{s}$ to $d_\mathrm{BART}$, and Transformer($\cdot$) is the application of a series of Transformer layers. 
 $\alpha$ is a scalar, in our case equal to $\sqrt{d_\mathrm{BART}}$, which is required to insure the input to BART is on the same scale as the embedding vectors BART was trained on. If we remove LN($\cdot$), $\mathbf{W}$ and $\alpha$, and set $d_\mathrm{s}=d_\mathrm{BART}$, we recover the method introduced by \citet{bart} for fine-tuning BART on MT.
\subsection{Extra Positional Embeddings}
\label{sec:extra}
We found empirically that the details of positional embedding vectors are important for good performance (see Table~\ref{tab:freeze-ablatate}), perhaps because of the need for the BART model to deal with different word order to that it was trained on. Transformer models normally have either learnable positional embedding vectors, or fixed sinusoidal positional embedding \cite{vaswani-attn} vectors $\mathbf{p}^t$, with
$\mathbf{p}_{i}^t = \mathrm{sin}( t / 10000^{i / (d_s / 2 - 1)})$,
if $0\leq i<d_s/2$, and $\mathbf{p}_{i}^t = \mathrm{cos}( t / 10000^{(i - (d_s / 2 - 1)) / (d_s / 2 - 1)})$ if $d_s / 2\leq i<d_s$, where $t$ indexes position and $i$ indexes dimension. 

Note that positional embedding are typically only added to the token embeddings. We use learnable positional embeddings at the embedding layer. But to get extra positional information, we optionally add fixed sinusoidal positional embedding to the input of each transformer layer in $\mathrm{IM(\cdot)}$, i.e.\ the input to layer $i$, $\mathbf{h}^i_t = \mathbf{o}^{i-1}_t + \mathbf{p}^t$, with $\mathbf{o}^{i-1}_t$ the previous layer output. This means the network has access to both \textit{learned positional embeddings} (only at the embedding layer), and \textit{fixed sinusoidal} ones at the input to each layer.
\begin{table*}[t]
\centering
\begin{tabular}{lcccr}
\toprule
{\bf Languages} & \textbf{Vi-En} & \textbf{Tuned Params (m)} \\
\hline
\textbf{(1)}: BART + InputModule (unfreeze all) &  9.5 & 374 \\
\textbf{(2)}: BART (frozen) + InputModule    & 27.9 & 26  \\
\textbf{(3)}: (2) + unfreeze layer-norm   & 28.4  & 26 \\
\midrule
(3) + sinusoidal positional embeddings &  18.3 & 26 \\
(1) + extra positional embeddings  &  22.0 & 26\\
\textbf{(4)}: (3) + extra positional embeddings &  29.0 & 26\\
\midrule
\textbf{(5)}: (3) + encoder adapters &  28.9 & 29 \\
(3) + decoder adapters &  28.3 & 29 \\
\midrule
\textbf{(6)}: (5) + extra positional embeddings  &  30.0 & 29 \\
\textbf{(7)}: (6) + GLU adapters &  30.5 & 29 \\

\hline
\end{tabular}
    \caption{Ablation study for various choices in the frozen BART method, with validation set BLEU score. We organise model settings by a number in brackets, (n), and define a new model configuration in bold as \textbf{(n)}:. We use `+' to indicate the addition of new model settings on top of the previous ones. Method \textbf{(2)} is similar to the method introduced by \citet{bart}. `+ sinusoidal positional embeddings' refers to adding sinusoidal positional embeddings to token embeddings, while `+ extra positional embeddings' refers to adding them within each transformer layer (see section~\ref{sec:extra}). `Tuned Params (m)' refers to the number of tunable parameters for each method in millions. Test set results are listed in Table~\ref{tab:mbart-en} (as `Frozen BART').}
\label{tab:freeze-ablatate}
\end{table*}

\begin{table}
\centering
\vskip 0.15in
\begin{tabular}{lcccr}
\toprule
{\bf Languages}  &{\bf It-En} &{\bf Si-En}   \\
\hline
(1): BART + InputModule + LN   & 34.1 & 5.1 \\
\hline
\textbf{(2)}: (1) + encoder adapters &  35.0 & 7.3\\
(1) + decoder adapters &  35.5 & 6.8\\
\hline
\textbf{(3)}: (2) + extra pos.\ embeddings  & 36.3 & 8.7 \\
\textbf{(4)}: (3) + GLU adapters  &  35.7 & 9.2 \\
\bottomrule
\end{tabular}
\caption{Further Ablation study for key settings of the frozen BART method, with validation set BLEU score. Test set results are listed in Table~\ref{tab:mbart-en} (as `Frozen BART').}
\label{tab:freeze-ablatate2}

\end{table}

\subsection{Within-Network Adapter Architecture}
\label{sec:adapter}
When freezing parts of a pre-trained model (either BART or mBART in our case), we may want to add flexibility by modifying the pre-trained model architecture. One approach is to use `adapters', introduced by \citet{houls, pals} which are newly-initialised neural network layers that can be `slotted in' to the layers of the pre-trained model. 

We only considered simple adapter architectures, essentially feed-forward networks, with one hidden layer, and a residual connection to the output. The dimension of the hidden layer can be much smaller than the model dimension to reduce computational cost and parameter count. We use one adapter per transformer layer, inserting them at the end of the layer \cite{pals, bapna-firat-2019-simple}. 
We use the following architectures, with $\mathbf{h}$ the hidden state of a particular token after the usual transformer layer, and  $\mathbf{h}_{\mathrm{out}}$ the hidden state of the token after the adapter layer:
\begin{equation}
\begin{split}
\mathbf{z} = \mathrm{gelu}(\mathbf{W}_d\mathbf{h}) & \\
   \mathbf{h}_{\mathrm{out}} = \mathrm{tanh}(\mathbf{W}_u\mathbf{z}) + \mathbf{h}
   \end{split}
\end{equation}
The $\mathrm{tanh}$ non-linearity helped with stability in early experiments, probably because it prevents the adapter output exploding by constraining it between -1 and 1. 

We also considered a version of the adapter based on the `gated linear unit' \citep[GLU;][]{gate} architecture:
\begin{equation}
\begin{split}
\mathbf{z} = 2 \sigma (\mathbf{W}_g\mathbf{h}) \odot \mathrm{gelu}(\mathbf{W}_d\mathbf{h}) & \\
   \mathbf{h}_{\mathrm{out}} = \mathrm{tanh}(\mathbf{W}_u\mathbf{z}) + \mathbf{h}.
   \end{split}
\end{equation}
We found the network was sensitive to changes in the magnitude of the hidden states the adapter produced, and therefore multiply the sigmoid gate by 2 so that it approximately leaves the magnitude of the hidden states unchanged.
\subsection{Freezing Details}
\paragraph{BART} We freeze all parameters of BART except the weights and biases of the layer-norm modules (following \citet{houls}), and additionally unfreeze the self-attention module of the first layer in the BART encoder, which is a small fraction of total BART parameters ($24 \cdot 2d_\mathrm{BART}$ from layer-norm parameters and $ 4 d_\mathrm{BART}^2$ from the self-attention module). We freeze BART token embeddings (used in the softmax layer).

\paragraph{mBART} In most of our experiments we unfreeze layer-norm parameters, positional and token embeddings, and either the entire encoder or decoder module (or the encoder and subsections of the decoder). We unfreeze the self-attention module of the first layer in the mBART encoder and decoder.

\begin{table*}[h]
\centering
\begin{tabular}{lccccccccr}
\toprule

{\bf Languages}  &  {\bf Vi-En$^\dagger$}  &  {\bf Vi-En }  &  {\bf It-En}  &  {\bf My-En}  &  {\bf Ne-En}  &  {\bf Si-En}  &  {\bf Cs-En}   &  {\bf Es-En} & \textbf{Pars (m)}\\
{\bf Size } &  110k  &  133k  &  250k  &  259k  &  564k  &  647k  &  11M  &  15M \\ 
\midrule
\textbf{(1)}: Freeze decoder  &  12.1 &  30.0  &  36.5  &  27.4  &  11.0  &  13.6  &  26.6  &  34.1 & 407 \\
Freeze encoder   &  12.0  &   29.7  &  36.6  &  25.2  &  8.8  & 12.3  &  25.6 &  33.8 & 457\\
\textbf{(2)}: (1) + adapters  &  12.2  &  30.0  &  36.7  &  27.7  &  10.8  &  14.2  &  27.4  &  34.4 & 410 \\
\midrule
(2) + ft enc-attn  &  12.3  &  30.6  &  37.0  &  29.0  &  11.4  &  14.9  &  27.0  &  35.1 & 461\\
(2) + ft self-attn  &  11.7  &  30.4  &  36.1  &  28.3  &  10.6  &  14.3  &  27.4  &  34.7 & 461\\
(2) + ft last 3 lyrs  &  12.1  &  30.6  &  36.6  &  28.1  &  11.5  &  14.7  &  27.6 &  34.9 & 461 \\
\midrule
\midrule
Test (random init)   &  8.1  &  23.6 &  31.7  &  23.3  &  7.6  &  7.2  &  22.0  &  29.0 & N/A \\
Test (frozen BART)   &  -   &  35.2  &  38.5  &  21.0  &  0.5  &  7.8  &  -  &  -  & 29 \\
Test (ft all)   &  14.1  &  \textbf{36.7}  &  \textbf{39.8}  &  \textbf{27.6}  &  14.1  &  \textbf{14.0}  &  29.2  &  \textbf{34.5} & 610 \\
Test (ft enc-attn)   &  \textbf{14.9}  &  \textbf{36.4}  &  39.4   &  \textbf{27.9}  &  \textbf{14.6}  &  \textbf{14.1}  &   \textbf{29.8}  &  \textbf{34.4} & 461 \\
\bottomrule
\end{tabular}
\caption{Validation BLEU score (unless stated otherwise) obtained by freezing various parts of the mBART and of adding adapters for \textit{Xx} $\rightarrow$ \textit{En}. `ft' refers to fine-tuning, i.e. unfreezing. Vi-En$^\dagger$ refers to a new parallel, `out-of-domain' dataset constructed similarly to the Flores \cite{flores} train sets (see section~\ref{sec:mbart-res}). `Test (frozen BART)' indicates results from English-only BART with the best performing method from Table~\ref{tab:freeze-ablatate2} or Table~\ref{tab:freeze-ablatate}. `Test (random init)' refers to training models (of various sizes) from scratch on the bitext for that language pair. `Pars (m)' refers to the number of tunable parameters for each method in millions (note token embeddings are tuned in every method and account for 256m parameters). Bold indicates the best test set score and all scores whose difference from the best is not statistically significant (with p-value less than 0.05).   (Statistical significance is computed via bootstrapping \cite{koehn-2004-statistical}.) }
\label{tab:mbart-en}
\end{table*}

\begin{table*}[h]
\centering
\begin{tabular}{lcccccccr}
\toprule

 & {\bf Vi-En}  & {\bf It-En} & {\bf My-En} & {\bf Ne-En} &  {\bf Si-En}  \\
\midrule
Freeze decoder (don't ft layer-norm)  &  29.6 & 35.1 & 26.6  & 10.3 & 13.1 \\
Freeze encoder (don't ft layer-norm)  &  29.4 & 36.1 & 24.1 & 8.7 & 12.1 \\

\bottomrule
\end{tabular}
\caption{Ablation study on improvement from fine-tuning layer-norm. Compare to the `Freeze decoder' and `Freeze encoder' methods in the first two rows of Table~\ref{tab:mbart-en}.}
\label{tab:mbart-ln}
\end{table*}

\begin{table*}[h]
\centering
\begin{tabular}{lcccccccr}
\toprule

{\bf Languages}  & {\bf En-Vi}   & {\bf En-It}  & {\bf En-My }  & {\bf  En-Ne}  & {\bf En-Si} & {\bf En-Cs}  & {\bf En-Es} & \textbf{Pars (m)} \\ 
\midrule
Freeze decoder    &   29.7  &  32.2  &  35.0  &  5.8  &  2.1  &  17.7 &  35.4 & 407 \\ 
\textbf{(1)}: Freeze encoder   &   30.1  &  31.5  &  36.0   &  5.3  &  3.7  &  16.5 &  35.0 & 457 \\ 
\textbf{(2)}: (1) + encoder adapters  &  30.3  &  32.3   &  36.9  &  5.4  &  4.2  &  16.6 &  35.3  & 461 \\ 
\midrule
\midrule
Test (ft all)   &  \textbf{35.4}  &  34.0  &  \textbf{36.9}  &  \textbf{7.4}  &  \textbf{3.3}  &  \textbf{18.0} &   \textbf{34.0} & 610  \\ 
Test (freeze enc. + adapters)   &  35.0  &  \textbf{34.3}  &  35.9  &  6.9  &  \textbf{3.3}  &  16.7 &  32.5  & 461 \\ 

\bottomrule
\end{tabular}
\caption{Validation BLEU score (unless stated otherwise) obtained by freezing various parts of the mBART and of adding adapters for for \textit{En} $\rightarrow$ \textit{Xx}. `Pars (m)' refers to the number of tunable parameters for each method in millions.}
\label{tab:mbart-xx}

\end{table*}
\section{Experimental Settings}
We use the fairseq \cite{ott2019fairseq} library for all experiments.
The final models are selected based on validation likelihood, except for multilingual fine-tuning where we evaluate the models after 10000 training steps.
We use beam-search with beam size $5$ for decoding, and evaluate all BLEU scores using SacreBLEU \cite{post-2018-call} 
\footnote{SacreBLEU signature: \texttt{BLEU+case.lc+lang.}
\texttt{[src-lang]-[tgt-lang]+numrefs.1+smooth}
\texttt{.exp+tok.13a+version.1.3.6}}. 
We use ISO 693-2 language codes in this work for convenience, and use the same parallel data as \citet{mbart}, both listed in listed in Table~\ref{tab:datastats} of the Appendix.

We fine-tune frozen BART and an Input Module on bilingual parallel text, feeding the source language into the Input Module. 
For mBART we feed the source language into the encoder, and use the same hyper-parameters as \citet{mbart}. When using adapters we use $0.1$ dropout in the adapter bottleneck layer ($\mathbf{z}$ in section~\ref{sec:adapter}), and a hidden dimension of either 128, or $\lfloor 2/3 \cdot 128 \rceil $ when using a gated linear unit adapter. 
We use the Adam \cite{kingma2014adam} optimizer.  Hyper-parameters are listed in Appendix~\ref{sec:hyper}, and we use the same hyper-parameter search space for frozen and non-frozen models.

\subsection{Multilingual MT}

We train with a very large effective batch size, training on 32 GPUs with a per-GPU batch size of 4096 tokens, meaning our total batch size is $N \cdot 32 \cdot 4096$ tokens, where $N$ is the number of language pairs. We evaluate our model after 10000 training steps (amounting to $N \cdot 10000$ forwards-backwards passes through the model).

\subsection{Vocabulary}

BART uses the GPT-2 tokenizer, which uses the BPE \cite{sennrich2016neural} approach (on the level of bytes, not characters). BART could technically take any Unicode string as input, however the BPE is learned on English text. When fine-tuning BART on machine translation we therefore learn a new subword vocabulary (using the sentencepiece \cite{kudo-richardson-2018-sentencepiece} library) on the source data from the fine-tuning dataset, and use a smaller vocabulary size of 5000, which empirically performs better for low-resource MT \cite{flores, revisiting}. We don't change the mBART tokenizer or vocabulary.

\section{Results and Discussion}
\subsection{Frozen BART}

Table~\ref{tab:freeze-ablatate} shows the effects of various choices we made in fine-tuning BART for MT. \textit{Freezing} is important: we see an 18.4 BLEU point improvement from fine-tuning a frozen BART model compared to fine-tuning an unfrozen BART (both with an Input Module; see section~\ref{sec:im}).

Adding extra flexibility with within-network adapters helps performance, especially when added to the BART \textit{encoder}. 
It is important to use \textit{learned positional embeddings} at the embedding layer in the Input Module, with an 10.1 BLEU score drop if we use fixed positional embeddings (at the embedding layer). We see consistent gains in Table~\ref{tab:freeze-ablatate} and Table~\ref{tab:freeze-ablatate2} by adding additional, fixed sinusoidal positional embeddings to the input of every transformer layer of the Input Module (see section~\ref{sec:extra}), even when using an unfrozen BART\@. The BART encoder `expects' English input, and it may be the Input Module with extra fixed embeddings can better account for the different word order in the input language. 
In the next section we compare to mBART and baselines.

\subsection{Frozen mBART}
\label{sec:mbart-res}

In Table~\ref{tab:mbart-en} and Table~\ref{tab:mbart-xx} we list results from freezing various parts of mBART\@. We get better performance than fine-tuning (`ft all' in Table~\ref{tab:mbart-en}) with our freeze decoder + fine-tune encoder-decoder attention method (`ft enc-attn' in Table~\ref{tab:mbart-en}) on Ne-En and Cs-En for \textit{Xx} $\rightarrow$ \textit{En}, and mostly similar results to the baseline otherwise. 

We believe a benefit to freezing, when fine-tuning on training data from a different domain to test data, will be avoiding specialising the pre-trained model to the fine-tuning train data domain. To test this we constructed a new Vi-En parallel dataset (Vi-En$^\dagger$ in Table~\ref{tab:mbart-en}) using the some of the same sources as the Flores \cite{flores} training data (the Si-En and Ne-En training sets used in this work), specifically GNOME/KDE/Ubuntu domain from the OPUS repository\footnote{http://opus.nlpl.eu/} and Bible translations from the bible-corpus\footnote{https://github.com/christos-c/bible-corpus/}, and use the same test and validation sets as the IWSLT15 Vi-En dataset. By constraining ourselves to this out-of-domain training set we see the largest gains out of the language pairs we considered over the fine-tuning baseline (0.9 BLEU).

We also consider the effect of the size of the fine-tuning dataset. If we constrain the training data to a random subset of 200k training examples from Ro-En (Table~\ref{tab:ro-en}), the `ft enc-attn' method outperforms simple fine-tuning. This effect generalises to an mBART variant that was pre-trained on only Ro and En monolingual data (using the same data as \citet{mbart}). Further results on Ro-En data are available in the Appendix, Table~\ref{tab:ro-en2}, and show similar trends to Table~\ref{tab:mbart-en}, with fine-tuning encoder-decoder attention the most important.

\begin{table*}[h]
\centering
\begin{tabular}{lccccr}
\toprule
 {\bf Model } & {\bf mBART }   & & {\bf En-Ro mBART } & &    \\
\midrule
{\bf Languages (Size)} &   {\bf Ro-En (608k)} & {\bf Ro-En (200k) } &{\bf Ro-En (608k)} & {\bf Ro-En (200k) }   \\
\midrule
Test (ft all)  & 37.8 & 36.4 & 38.5 & 37.7 \\
Test (ft enc-attn)  & 37.8 & 36.8 & 38.1 & 37.9 \\
\bottomrule
\end{tabular}
\caption{Validation set BLEU (unless stated otherwise) comparing freezing various parts of mBART and En-Ro mBART (pre-trained only on En and Ro data), fine-tuned on \textit{Ro} $\rightarrow$ \textit{En} parallel data. `ft' referes to fine-tuning, i.e. unfreezing. `Ro-En (200k)' refers to a random subset of the Ro-En training data of size 200k.}
\label{tab:ro-en}
\end{table*}

\begin{table*}[h]
\centering
\begin{tabular}{lcccccccccccccccr}
\toprule

{\bf Src. Lang.} & {\bf   Ru} &{\bf  Fr} &{\bf  De} & {\bf Zh} &{\bf  Es} &{\bf  Cs} &{\bf  Lv} &{\bf  Fi} &{\bf  Lt} &{\bf  Et} &{\bf  Hi} &{\bf  Si}  \\
{\bf Size} & 32M& 29M & 28M &  25M &  15M & 11M & 4.5M & 2.7M & 2.1M  &  1.9M &  788k &  647k  \\
\midrule
Finetune all  & {\bf 33.6} & {\bf 39.0} & {\bf 33.1}& {\bf\underline{20.2}} & \underline{33.7} & {\bf 29.9} & {\bf 21.1} & {\bf \underline{29.0}} & {\bf 22.8} & {\bf 28.6} & {\bf 25.4}  & {\bf 16.9}\\
Ft enc-attn  & {\bf 33.4} & 38.2 & 32.6 & {\bf\underline{20.2}} & {\bf \underline{34.0}}& 29.7 & {\bf 20.8} & {\bf \underline{29.1}} & {\bf 22.7} & {\bf 28.3} & 25.1 & {\bf 16.7} & \\

\midrule
\midrule

{\bf Src. Lang.} &{\bf  Ro}  & {\bf Ne} &{\bf  My} &{\bf  Ar} &{\bf  It} & {\bf Nl} &{\bf  Ko} &{\bf  Ja} &{\bf  Tr}  &{\bf  Vi} &{\bf  Kk} &{\bf  Gu}  \\
{\bf Size} & 612k & 563k & 259k &  251k& 251k &  237k &230k &223k & 207k &  133k& 91k & 12k  \\
\midrule
Finetune all  & {\bf 37.8} & {\bf \underline{20.7}} & {\bf 31.0 }& {\bf 37.0} &{\bf 39.6} & {\bf 43.3} &{\bf 25.0}& {\bf \underline{18.7}} & {\bf 24.0} & {\bf \underline{37.4}}& \underline{14.6}& {\bf 18.7} \\
Ft enc-attn  &{\bf 37.9}& {\bf \underline{20.8}} & 30.5 & {\bf 36.9} & {\bf 39.3} & {\bf 43.0}& 24.2& {\bf \underline{18.8}}& 23.7& {\bf \underline{37.5}}& {\bf \underline{15.0}}& {\bf 18.3} \\


\bottomrule
\end{tabular}
\caption{Test set BLEU score on many-to-one (\textit{Xx} $\rightarrow$ \textit{En}) multilingual MT with a simple round-robin training schedule.  `Ft enc-attn' refers to fine-tuning the encoder, and fine-tuning the encoder-decoder attention module in every decoder layer, leaving the other decoder sub-modules frozen. The `Ft enc-attn' model setting uses adapter modules in the decoder to increase flexibility after freezing parameters. Bold indicates the best score
and all scores whose difference from the best is not statistically significant (with p-value less than 0.05). For clarity we underline language pairs where the `Ft enc-attn' method matches or outperforms naive fine-tuning.}
\label{tab:mbart-multi}

\end{table*}

Table~\ref{tab:mbart-en} shows the relative performance of frozen BART, frozen mBART and baselines. Fine-tuning mBART gave consistently better results than frozen BART\, especially for distantly related languages. For Si, Ne and My the performance of frozen BART is roughly on par with a randomly initialised model (or much worse in the case of Ne-En). The parallel data for these languages is often lower quality, and the BART system has to learn about the non-English language from noisy or out-of-domain text (e.g.\ text from the Ubuntu manual for the En-Ne pair). For Vi and It, we have high quality parallel data, and the frozen BART method is only approximately 1.5 BLEU points behind the best mBART results.  
We note mBART was trained on more English data than BART, and with different noising function hyper-parameters.

\subsection{What Should be Unfrozen?}
\label{sec:what}

\paragraph{Layer-Norm} 

We find large benefits to simply fine-tuning the weights and biases of the pre-trained layer-norm weights (recall that after normalisation, the layer-norm module multiplies each hidden dimension by a weight and adds a bias); this was observed in the setting of BERT by \citet{houls}. This gains e.g. 0.5 BLEU for frozen BART (see Table~\ref{tab:freeze-ablatate}) and an average of 0.8 BLEU across five languages for mBART (see Table~\ref{tab:mbart-ln} compared to Table~\ref{tab:mbart-en}). Since these weights and biases are only 2$d$ parameters per layer-norm, where $d$ is the model dimension. This is parameter-efficient, with adding more parameters with `Adapters' on top of unfrozen layer-norm providing a smaller improvement.
\paragraph{Encoder vs Decoder}
For the \textit{Xx} $\rightarrow$ \textit{En} direction (Table~\ref{tab:mbart-en}) we can see that freezing the decoder always performs better than freezing the encoder (except for It-En where they perform roughly the same.) 
For the \textit{En} $\rightarrow$ \textit{Xx} direction (Table~\ref{tab:mbart-xx}) we see slightly weaker evidence for the opposite trend, with the decoder more useful to fine-tune; but for the high resource languages Es and Cs freezing the decoder works better. 
There is more English data in mBART pre-training than data in other languages, which may account for better results with a frozen encoder (when English is the source language) or decoder (when English is the target language). Adding flexibility with adapters in the frozen layers improves performance in all languages and directions, except for Ne$\rightarrow$En. 

We explore more fine-grained unfreezing for the \textit{Xx} $\rightarrow$ \textit{En} direction (Table~\ref{tab:mbart-en}). We fine-tuned three equally sized subsets of the decoder: the encoder-decoder attention layers (approx.\ $12 \cdot 4 d_\mathrm{BART}^2$ parameters), the self-attention layers in the decoder (approx.\ $12 \cdot 4 d_\mathrm{BART}^2$ parameters), or the entire last three layers of the decoder (approx.\ $3 \cdot 16 d_\mathrm{BART}^2$ parameters). We observe that fine-tuning the encoder-decoder attention performed well (note the last three layers include three encoder-decoder attention layers), with fine-tuning self-attention the least useful. We hypothesize that the pre-training task of mBART (reconstructing noisy monolingual sentences) does not help with teaching the encoder-decoder attention to align source and target text of different languages.

\subsection{Memory Cost}
\label{sec:memory}
\begin{table}[h]
\centering
\begin{tabular}{lcr}
\toprule
 & {\bf Tokens per GPU}   \\
\midrule
Finetune all & 2304\\
\textbf{(1)}: Freeze decoder   & 4096  \\
Freeze encoder   &  3584 \\
\textbf{(2)}: (1) + decoder adapters &  4096 \\

(2) + ft enc-attn & 3328  \\

\bottomrule
\end{tabular}
\caption{Maximum number of tokens that would fit on one NVIDIA Volta GPU when fine-tuning mBART on the En-Vi training set. We evaluated batch sizes in increments of 256 tokens.}
\label{tab:memory}
\end{table}

Freezing parameters means we no longer need to allocate memory to storing their gradients. We will obtain additional memory savings when using an optimizer that stores various other quantities (i.e.\ the Adam optimizer stores running averages of the first and second moments of gradients.). The memory savings allow for roughly 45-75\% larger batches for the methods we consider in this work (see Table~\ref{tab:memory} for our mBART methods), but for larger pre-trained models the proportion of GPU memory freed up by freezing will increase. At inference time we no longer require gradients and we have the same memory cost.

\subsection{Multilingual Fine-tuning of mBART}
\label{sec:multi}
We explore freezing parts of the mBART model when fine-tuning on a challenging multilingual MT task. Table~\ref{tab:mbart-multi} lists results from a naive fine-tuning baseline, and results from freezing most of the decoder but unfreezing the encoder-decoder attention (when freezing we use GLU adapters in the decoder, see section~\ref{sec:adapter}). Freezing parameters hurts performance on some language pairs, and since freezing removes flexibility from the model and we have to adapt to 25 different directions this is perhaps not surprising. The language pairs where we match or improve on the baseline are Zh, Es, Fi, Ne, Ja, Vi and Kk. These are mostly (five out of seven) non-European languages, and distantly related to En. However since most of these results are not statistically significant further study is needed to verify this. Note we see a clear benefit over bilingual fine-tuning for some language pairs (e.g.\ compare our best Ne result from Table~\ref{tab:mbart-en}, 14.6 BLEU vs.\ 20.8 BLEU for multilingual fine-tuning). We leave to future work a more thorough investigation of the multilingual MT setting. 

\section{Conclusion}

We recommend: For a language with high quality parallel data but without a pre-trained model trained on monolingual data from that language, using a frozen (English-only) BART model with additional parameters at the source side (the `input module') improves performance over a randomly initialised baseline. For this approach it is important to freeze the pre-trained model. We also give the model both learned positional embeddings at the embedding layer, and fixed sinusoidal positional embeddings at each layer of the input module. 

For a multilingual pre-trained model, we found performance improvements on some (mostly distantly related) languages for  multilingual many-to-one fine-tuning. For bilingual \textit{En} $\rightarrow$ \textit{Xx} fine-tuning we did not see any improvement, although the performance drops are small, and by freezing parameters we need less memory at training time compared to fine-tuning. For \textit{Xx} $\rightarrow$ \textit{En} bilingual fine-tuning it is important to unfreeze the encoder-decoder attention, and keep the rest of the decoder frozen. This can improve on simple fine-tuning, especially for distantly-related language pairs or those with out-of-domain training data.

We recommend fine-tuning layer-norm parameters as a parameter-efficient complement to adapter layers. For our mBART experiments we found it was necessary to fine-tune the token embeddings, which correspond to a large number of parameters, and future work could remove this cost by working out a subset of the vocabulary to fine-tune, or another method. 

\section*{Acknowledgments}
We’d like to thank James Cross, Mike Lewis, Naman Goyal, Jiatao Gu, Iain Murray, Yuqing Tang and Luke Zettlemoyer for useful discussion. We also thank our colleagues at FAIR and FAIAR for valuable feedback.

\bibliography{anthology,eacl2021,emnlp2020,extrarefs}
\bibliographystyle{acl_natbib}

\appendix
\section{Additional Ablation Study}
In Table~\ref{tab:mbart-ln2} we reproduce Table ~\ref{tab:mbart-ln} of the main paper with more context to study the effect of unfreezing layer-norm parameters when fine-tuning mBART\@. Across all language pairs we see improvements from fine-tuning layer norm parameters over not fine-tuning them, and additional, smaller, improvements from adding adapters, indicating both forms of adding flexibility are useful. In Table~\ref{tab:ro-en2} we present additional results on the Ro-En pre-trained model (see section 3.2 of the main body).
\begin{table*}[h]
\centering
\begin{tabular}{lcccccccr}
\toprule

 & {\bf Vi-En}  & {\bf It-En} & {\bf My-En}  & {\bf Ne-En} & {\bf Si-En} \\
\midrule
Freeze decoder   &  26.6 & 35.1 & 26.6 & 10.3 & 13.1 \\
Freeze encoder  &  29.4 & 36.1 & 24.1 & 8.7 & 12.1  \\
\midrule
\textbf{(1)}: Freeze decoder + ft layer norm  &  30.0 & 36.5 & 27.4& 11.0 & 13.6  \\
Freeze encoder + ft layer norm &  29.7 & 36.6 & 25.2 & 8.8 & 12.3 \\
(1) + decoder adapters & 30.0 & 36.7 &  27.2 & 10.8 & 14.2 \\

\bottomrule
\end{tabular}
\caption{Validation BLEU score (unless stated otherwise) obtained by fine-tuning layer-norm parameters and of adding adapters for mBART, for \textit{Xx} $\rightarrow$ \textit{En}. `ft' refers to fine-tuning, i.e. unfreezing. Note we are simply reproducing rows from Table~\ref{tab:mbart-en} and Table~\ref{tab:mbart-ln} of the main paper for ease of comparison. }
\label{tab:mbart-ln2}
\end{table*}

\begin{table*}[h]
\centering
\begin{tabular}{lccccr}
\toprule
 & {\bf mBART }   & & {\bf En-Ro mBART } & &    \\
\midrule
 &   {\bf Ro-En (608k)} & {\bf Ro-En (200k) } &{\bf Ro-En (608k)} & {\bf Ro-En (200k)} \\
 \midrule
\textbf{(1)}: Freeze decoder   & 38.8 & 37.9 & 40.4 & 39.9 \\
Freeze encoder   & 39.1 & 38.3 & 40.0 & 39.2 \\
\textbf{(2)}: (1) + decoder adapters & 39.3 & 38.0 & 40.6 & 40.0 \\
\midrule
(1) + ft enc-attn & 39.8 & 39.0 & 40.5 & 40.5 \\
(1) + ft self-attn & 39.6 & 38.3 & 40.4 & 40.1 \\
(1) + ft last 3 lyrs & 39.6 & 38.6 & 40.5 & 40.3 \\
\midrule
\midrule
Test (ft enc-dec)  & 37.8 & 36.8 & 38.1 & 37.9 \\
Test (ft all)  & 37.8 & 36.4 & 38.5 & 37.7 \\
\bottomrule
\end{tabular}
\caption{Validation set BLEU (unless stated otherwise) comparing freezing various parts of mBART and En-Ro mBART (pre-trained only on En and Ro data rather than 25 languages), fine-tuned on \textit{Ro} $\rightarrow$ \textit{En} parallel data. `ft' refers to fine-tuning, i.e. unfreezing. `Ro-En (200k)' refers to a random subset of the Ro-En training data of size 200k.}
\label{tab:ro-en2}
\end{table*}
\section{Fine-tuning Hyper-parameters}
\label{sec:hyper}
For all experiments with bilingual datasets we use a batch size of 2048$\times$16 tokens, i.e.\ 2048 tokens per GPU and 16 GPUs (we investigate larger batch sizes for frozen models only to test GPU memory usage, and do not evaluate models trained with larger batch sizes). Ranking of hyper-parameters was done by validation set BLEU score.
\paragraph{Frozen BART}

We train with $0.3$ dropout for the frozen BART parameters, and $0.2$ dropout for the Input Module parameters, $0.1$ label smoothing, $0.2$ dropout for the self-attention scores in the Input Module, $5000$ warm-up steps, and $7\mathrm{e}{-4}$ maximum learning rate. We performed a grid search over learning rates in \{$7\mathrm{e}{-4}$, $5\mathrm{e}{-4}$, $3\mathrm{e}{-4}$\}, dropout for Input Module parameters in \{$0.2$, $0.1$\}, and dropout for self-attention scores in \{$0.2$, $0.1$\}.
We train for a maximum of $50$K training updates for all low and medium resource pairs and $100$K for high resource pairs (which takes roughly 8 hours and 16 hours respectively). 

\paragraph{Frozen mBART}

We train with $0.3$ dropout, $0.2$ label smoothing, $2500$ warm-up steps, and $3\mathrm{e}{-5}$ maximum learning rate. We did not search over hyper-parameters, simply re-using those of \citet{mbart}. Despite the adapter parameters being randomly initialised, the small learning rate did not affect performance (we performed a small sweep of larger learning rates and found only marginal gains, and so kept the same settings for simplicity).
 We use a maximum of $40$K training updates for all low and medium resource pairs and $100$K for high resource pairs (Es and Cs in our case), this takes roughly 12 hours and 30 hours respectively.

\paragraph{Multi-lingual MT}

We train with $0.3$ dropout, $0.1$ dropout for self-attention scores, $4000$ warm-up steps, and $1\mathrm{e}{-4}$ maximum learning rate. 

\paragraph{Out-of-domain Vi-En Baseline}

To train a randomly initialised baseline for the out-of-domain Vi-En data (Vi-En$^\dagger$ in Table~\ref{tab:mbart-en} of the main body) we used the same model architecture and training settings as those of \citet{flores} use for training MT systems on similar data (but with Si or Ne source language). Specifically a seq2seq transformer with 5 encoder and decoder layers, hidden dimension 512. shared embeddings between the input and softmax layers, and strong regularisation (e.g.\ 0.4 dropout on hidden states, 0.2 dropout on attention scores, 0.2 label smoothing). We learn a BPE vocabulary (joint across source and target data) of size 5000 on the training data. For full details of hyper-parameters we refer the reader to \citet{flores} and the associated GitHub repository\footnote{https://github.com/facebookresearch/flores}.

\section{Pre-training Languages}
\label{sec:langs}
\begin{table}[t]
\begin{center}
\small
\begin{tabular}[b]{llrrr}
\toprule
\textbf{Code} & 
\textbf{Language} & 
\textbf{Tokens(M)} & \textbf{Size(GB)}  & \textbf{Parallel data source}  \\
{\bf En }& English & 55608 & 300.8 &  \\
{\bf Ru }& Russian & 23408 & 278.0 & WMT19\\
{\bf Vi }& Vietnamese & 24757 & 137.3 &  IWSLT15\\
{\bf Ja }& Japanese & 530 (*) & 69.3  & IWSLT17\\
{\bf De}& German & 10297 & 66.6 & WMT19\\
{\bf Ro }& Romanian & 10354 & 61.4 & WMT16 \\
{\bf Fr }& French & 9780 & 56.8 & WMT19 \\
{\bf Fi }& Finnish & 6730 & 54.3  & WMT17 \\
{\bf Ko }& Korean & 5644 & 54.2  & IWSLT17\\
{\bf Es }& Spanish & 9374 & 53.3 & WMT19\\
{\bf Zh } & Chinese (Sim) & 259 (*) & 46.9 & WMT19\\
{\bf It }& Italian & 4983 & 30.2 &  IWSLT17\\
{\bf Nl }& Dutch & 5025 & 29.3  & IWSLT17\\
{\bf Ar }& Arabic & 2869 & 28.0  & IWSLT17 \\
{\bf Tr }& Turkish & 2736 & 20.9  & IWSLT17\\
{\bf Hi }& Hindi & 1715 & 20.2 & ITTB \\
{\bf Cs }& Czech & 2498 & 16.3 & WMT19 \\
{\bf Lt }& Lithuanian & 1835 & 13.7 & WMT19 \\
{\bf Lv }& Latvian & 1198 & 8.8  & WMT17\\
{\bf Kk }& Kazakh & 476 & 6.4  & WMT19\\
{\bf Et }& Estonian & 843 & 6.1  & WMT18\\
{\bf Ne }& Nepali & 237 & 3.8 & FLoRes\\
{\bf Si }& Sinhala & 243 & 3.6 & FLoRes \\
{\bf Gu }& Gujarati & 140 & 1.9 & WMT19\\
{\bf My }& Burmese & 56 & 1.6 & WAT19\\
\bottomrule
\end{tabular}
\caption{\textbf{Languages and Statistics of the CC25 Corpus.} A list of the 25 languages used in mBART pre-training ranked with monolingual corpus size. 
(*) The Chinese and Japanese corpora are not segmented, so the token counts here are sentence counts.}
\label{tab:datastats}
\end{center}
\end{table}

We reproduce in Table~\ref{tab:datastats} the details from \citet{mbart} of the size of each pre-training language corpus for mBART\@. 

\end{document}